%% file: dropout_sampling.tex
\documentclass[letterpaper, 10 pt, conference]{ieeeconf}  

\IEEEoverridecommandlockouts                              

\usepackage{graphicx} 
\usepackage{amsmath} 
\usepackage{amssymb}  
\usepackage{booktabs}
\usepackage{cite}

\usepackage{xcolor}

\input{mathstuff}

\usepackage{flushend}

\title{\LARGE \bf
Dropout Sampling for Robust Object Detection in Open-Set Conditions
}

\author{Dimity Miller, Lachlan Nicholson, Feras Dayoub, Niko S\"underhauf
\thanks{This research was supported by the Australian Research Council Centre
of Excellence for Robotic Vision, project number CE140100016.}
\thanks{The authors are with the Australian Centre for Robotic Vision at Queensland University of Technology (QUT),
Brisbane, QLD 4001, Australia.  Contact: {\tt\small dimity.miller@hdr.qut.edu.au}}%
}

\begin{document}

\maketitle
\thispagestyle{empty}
\pagestyle{empty}

\begin{abstract}
Dropout Variational Inference, or Dropout Sampling, has been recently proposed as an approximation technique for Bayesian Deep Learning and evaluated for image classification and regression tasks. This paper investigates the utility of Dropout Sampling for object detection for the first time. We demonstrate how label uncertainty can be extracted from a state-of-the-art object detection system via Dropout Sampling. We evaluate this approach on a large synthetic dataset of 30,000 images, and a real-world dataset captured by a mobile robot in a versatile campus environment. We show that this uncertainty can be utilized to increase object detection performance under the open-set conditions that are typically encountered in robotic vision. A Dropout Sampling network is shown to achieve a 12.3\% increase in recall (for the same precision score as a standard network) and a 15.1\% increase in precision (for the same recall score as the standard network). 
\end{abstract}

\section{Introduction}\label{sec:intro}
Visual object detection has made immense progress over the past years thanks to advances in deep learning and convolutional networks \cite{Liu16,YOLO9000,Ren15}. Despite this progress, operating in open-set conditions, where new objects that were not seen during training are encountered \cite{scheirer2013toward, scheirer2014probability}, remains one of the biggest current challenges in visual object detection.

Robots that have to operate in ever-changing, uncontrolled real-world environments commonly encounter open-set conditions and have to cope with new object classes that were not part of the training set of their vision system.

This scenario is very different to how current visual object detection systems are evaluated. Typically one large dataset is split into a training and testing subset that is used for evaluation. As a result, both sets share the same characteristics and contain the same object classes. This is commonly referred to as operating under closed-set conditions, where all objects seen during testing are also known during training. It was shown in~\cite{Torralba11} that top performing object classification and recognition systems suffer a major drop in performance when tested using samples taken from outside their ``universe'', i.e tested on images taken from outside the particular dataset used for training and testing.

Solving the open-set object detection problem is of paramount importance for the successful deployment of learning-based systems on board of mobile robots. A robot that acts based on the output of an unreliable machine learning system can potentially have serious repercussions. 

One way to handle the open-set problem is to utilize the uncertainty of the model predictions to reject predictions with low confidence. An approach to this uncertainty estimation has been developed by the use of a technique called Dropout Sampling as an approximation to Bayesian inference over the parameters of deep neural networks~\cite{gal2016dropout}. Consequently, this technique has been used for uncertainty estimation in image classification and regression tasks~\cite{kendall2016bayesian,kendall2016modelling} but has not yet been utilized for object detection.

The objective of this paper is to extend the concept of Dropout Sampling to object \emph{detection} for the first time. We achieve this by evaluating a Bayesian object detection system on a large synthetic and a real-world dataset and demonstrating that the estimated label uncertainty can be utilized to increase object detection performance under open-set conditions.

\begin{figure}[t]
    \centering
    \includegraphics[width=0.7\linewidth]{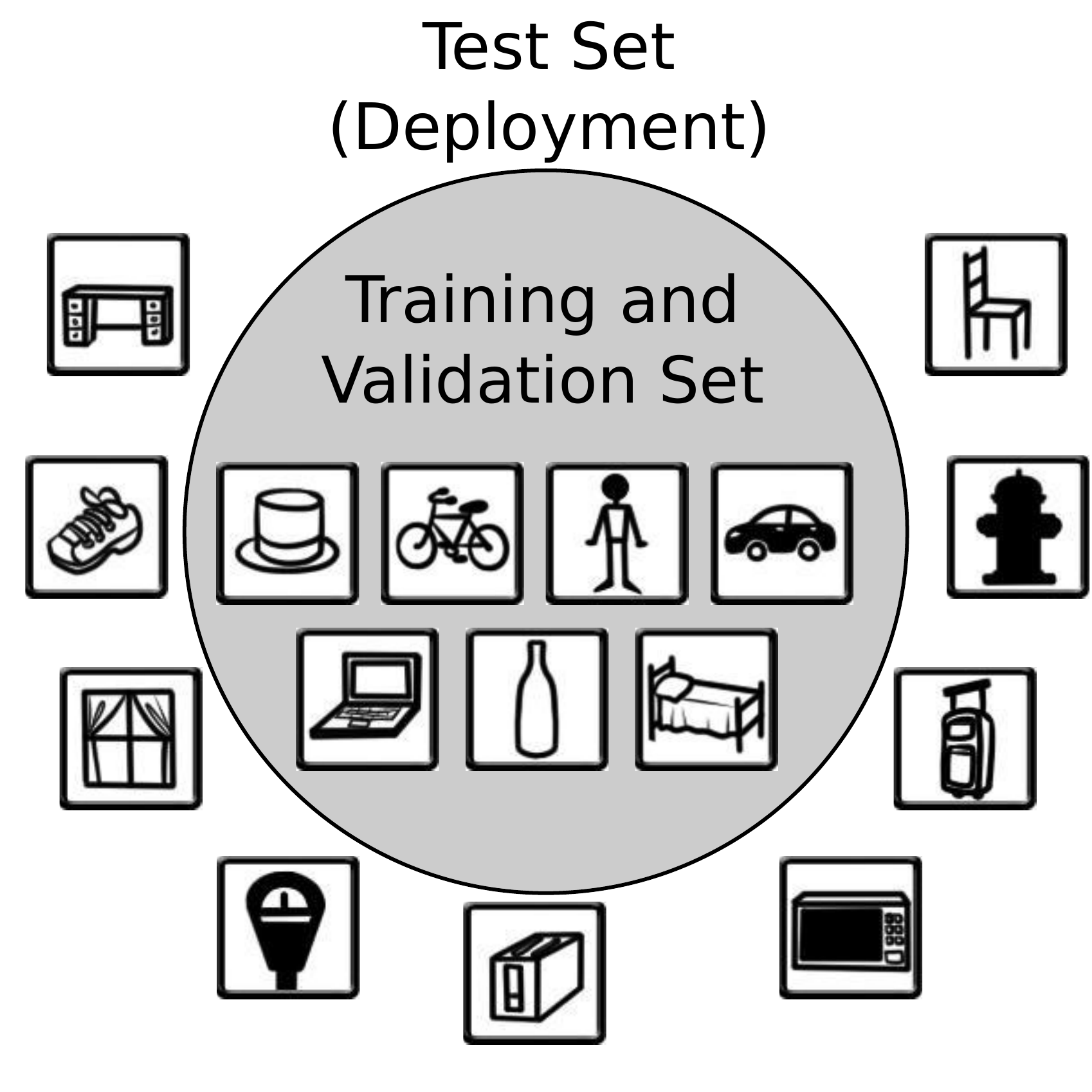}
    \caption{The Open-Set problem. Training of an object detection system is performed on a closed set of known classes. In typical computer vision benchmarks such as COCO \cite{Lin14} or ILSVRC \cite{Russakovsky15} the test set is identical to the training set, i.e. there are no new classes in the test set. In stark contrast, robots operating in the real world in uncontrolled environments commonly encounter many objects of previously unseen classes. Icons in this image have been taken from the COCO dataset website (http://cocodataset.org/\#explore).
    }
    \label{fig:openset}
\end{figure}

The remainder of the paper is structured as follows; Section~\ref{sec:related} discusses the related work with Section~\ref{sec:approach} presenting our proposed approach to obtaining uncertainty estimation for object detection. Section~\ref{sec:metrics} describes the evaluation metrics and the datasets used. Section~\ref{sec:results} describes the experimental evaluation and the results. Finally, Section~\ref{sec:conc} draws conclusions and discusses future research.
\section{Related Work}\label{sec:related}
\subsection{Visual Object Detection}
Visual object detection is the process of finding all instances of known object classes in an image and accurately localizing it using a tight bounding box.

Current state-of-the-art visual object detection systems are dominated by deep neural networks. The first breakthrough was in 2014 by R-CNN~\cite{Girshick14} which used cropped and resized regions from an input image using a regions proposals as an input to a deep convolutional neural network classifier, AlexNet~\cite{Krizhevsky12}, in order to localize all known objects. Later and in order to improve the speed of the training and testing stages of R-CNN, Faster R-CNN~\cite{Ren15} integrated the process of region proposal generation as a branch in the network itself. Recently, Single shot multibox detector (SSD)~\cite{Liu16} took the idea further and unified the detection and proposal generation into one branch in the network. This enabled the detector to consider different image regions of different sizes and resolutions. 

Although theses networks are performing increasingly well under \emph{closed-set} conditions, they suffer performance loss when evaluated using images from outside their corresponding development datasets (i.e a similar setup to \emph{open-set} conditions) as shown in~\cite{Torralba11}. 

\subsection{Open-set Object Detection}
Open-set conditions is defined as the evaluation of a system where novel classes are seen in testing that were not present during training. 
As defined in \cite{scheirer2014probability}, there exists three categories of classes:
\begin{enumerate}
    \item \emph{Known classes, i.e.} the classes with distinctly labeled positive training examples,
    \item \emph{Known unknown classes, i.e.} labeled negative examples, not necessarily grouped into meaningful categories,
    \item \emph{Unknown unknown classes, i.e.} classes unseen during training. 
\end{enumerate}
Although some modern object detectors are trained to detect ``background'' classes (\emph{known unknown classes}) and distinguish them from \emph{known classes}, it is not possible to \emph{train} a system to detect and discriminate against \emph{unknown unknown classes}. 

The problem with deploying models trained under closed-set assumptions into open-set environments is that the network is forced to choose a class label from one of the \emph{known} classes, and in many cases, classifies the unknown object as a known class with high confidence ~\cite{bendale2016towards}. 

Current attempts at improving open-set performance of machine learning systems have focused on formally accounting for \emph{unknown unknowns}~\cite{scheirer2013toward, scheirer2014probability, rudd2017extreme} by identifying and rejecting classes not encountered during training based on an estimate of the uncertainty in the network predictions.

\subsection{Bayesian Deep Learning}
One way to obtain an estimate of uncertainty is by using Bayesian Neural Networks (BNNs)~\cite{mackay1992practical, neal1995bayesian}. Commonly, variational inference has been used to obtain approximations for BNNs as shown in~\cite{paisley2012variational, kingma2013auto, rezende2014stochastic, titsias2014doubly, hoffman2013stochastic}. However, the practical applicability of these methods is hindered by increased training difficulty and computational cost. 

In 2015, Gal and Ghahramani \cite{gal2016dropout} proposed Dropout Variational Inference as a tractable approximation to BNNs that provides a measure of uncertainty for a models confidence scores while remaining computationally feasible. This made it possible for any deep neural network to become Bayesian by simply enabling the dropout layers during testing, as opposed to standard practice where dropout layers are only used during training.

Recently, in~\cite{kendall2016bayesian} and \cite{kendall2016modelling}, dropout sampling was used for uncertainty estimates on regression and image classification tasks in order to improve performance. In this paper, we extend the use of this technique to visual object detection, where multiple objects in a scene are localized and classified. We then evaluate the effect of this technique on object detection performance under open-set conditions typical to robot vision tasks.
\section{Object Detection -- A Bayesian Perspective}\label{sec:approach}
We start by giving a short overview on how Dropout Sampling is used to perform tractable variational inference in classification and recognition tasks. We then present our approach to extending this technique to object \emph{detection}. 

\subsection{Dropout Sampling for Classification and Recognition}
The idea behind Bayesian Neural Networks is to model the network's weights $\vW$ as a distribution $p(\vW | \vT)$ conditioned on the training data $\vT$, instead of a deterministic variable. By placing a prior over the weights, e.g. $\vW \sim \normal{0}{\vI}$, the network training can be interpreted as determining a plausible set of weights $\vW$ by evaluating the posterior over the weights given the training data: $p(\vW | \vT)$~ \cite{kendall2017uncertainties}. Evaluating this posterior however is not tractable without approximation techniques. 

Kendall and Gal \cite{kendall2017uncertainties} showed that for \emph{recognition} or classification tasks, Dropout Variational Inference allows the approximation of the class probability $p(y | \cI, \vT)$ given an image $\cI$ and the training data $\vT$ by performing multiple forward passes through the network with Dropout enabled, and averaging over the obtained Softmax scores $\vs_i$:
\begin{equation}
    p(y | \cI, \vT) =\int p(y | \cI, \vW)\cdot p(\vW|\vT) d\vW \approx \frac{1}{n} \sum_{i=1}^n \vs_i
    \label{eq:do_sample}
\end{equation}
This Dropout Sampling technique essentially \emph{samples} $n$ model weights $\widetilde\vW_i$ from the otherwise intractable posterior $p(\vW | \vT)$.

In the above example, $p(y | \cI, \vT)$ is a probability vector $\vq$ over all class labels. The uncertainty of the network in its classification is captured by the entropy $H(\vq) = -\sum_i q_i \cdot \log q_i$. This technique of estimating uncertainty with Dropout Sampling has been successfully applied to various classification and regression tasks \cite{gal2016dropout,kendall2016modelling,kendall2016bayesian,kendall2017uncertainties}.

\subsection{Object Detection with Dropout Sampling}
In contrast to image classification or recognition that reports a single label distribution for what is considered the most prominent object in an image, object \emph{detection} is concerned with estimating a bounding box alongside a label distribution for multiple objects in a scene. We extend the concept of Dropout Sampling as a means to perform tractable variational inference from image recognition to object detection.

To do this, we employ the same Dropout Sampling approximation as proposed by \cite{gal2016dropout} to sample from the distribution of weights $p(\vW | \vT)$. This time however, $\vW$ are the learned weights of a \emph{detection} network, such as SSD~\cite{Liu16}. 

SSD is based on the VGG-16 network architecture \cite{simonyan2014SSD} that consists of 13 convolutional layers and 3 fully connected layers. This base network is trained with Dropout layers inserted after the first and second fully connected layers. Normally, these Dropout layers would not be active during testing, but we keep them enabled to perform the Dropout Sampling. Every forward pass through the network therefore corresponds to performing inference with different network $\widetilde\vW$ approximately sampled from $p(\vW|\vT)$. 

\subsection{Partitioning Detections into Observations}
\label{sec:partition}
\begin{figure*}[t]
    \centering
    \includegraphics[width=0.3\linewidth]{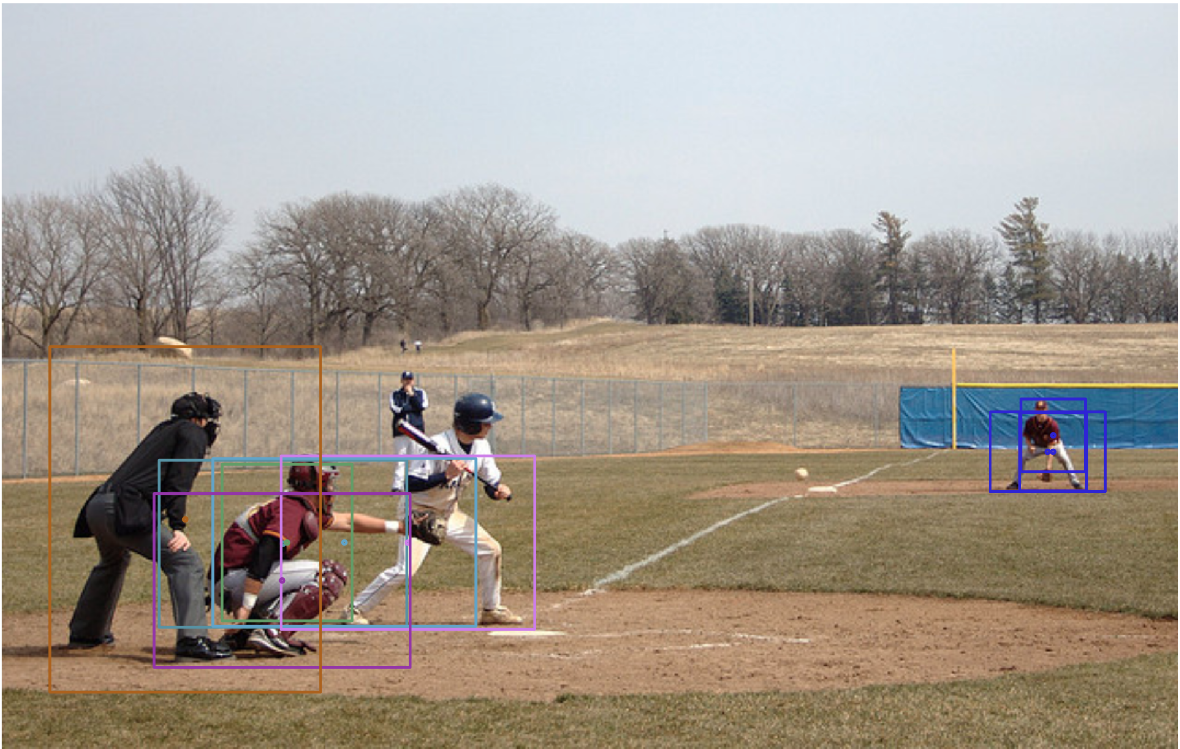}
    \includegraphics[width=0.3\linewidth]{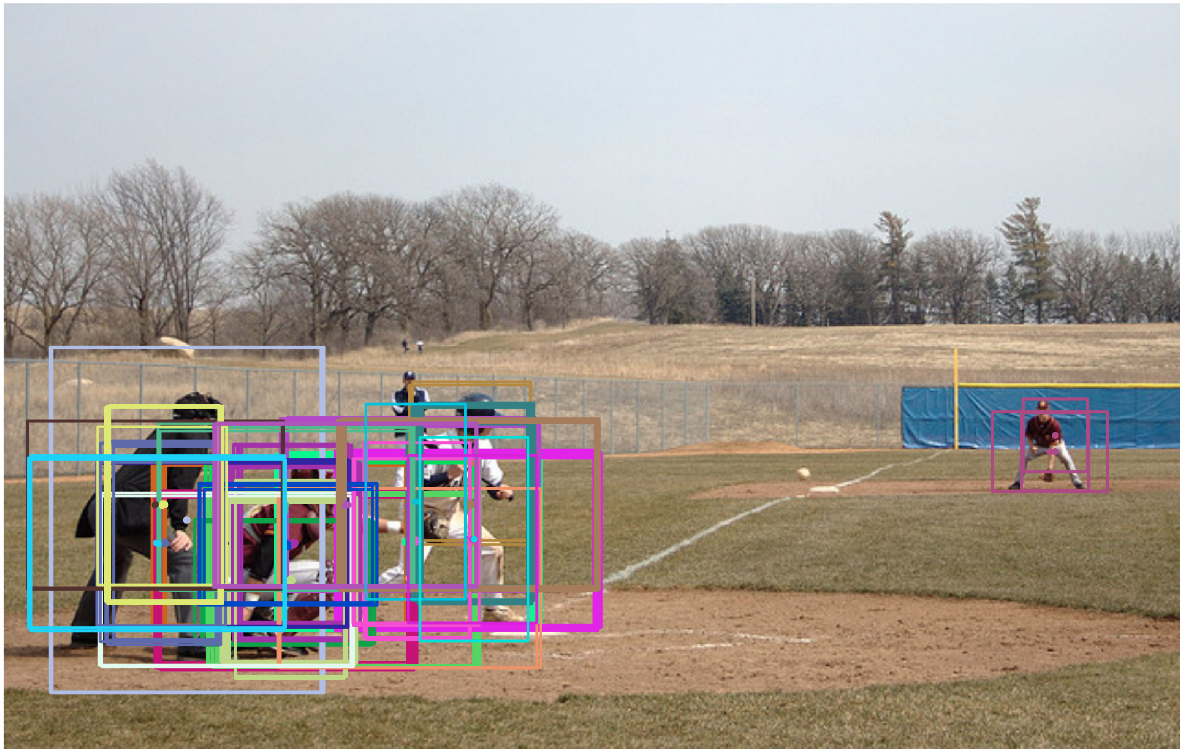}
    \includegraphics[width=0.3\linewidth]{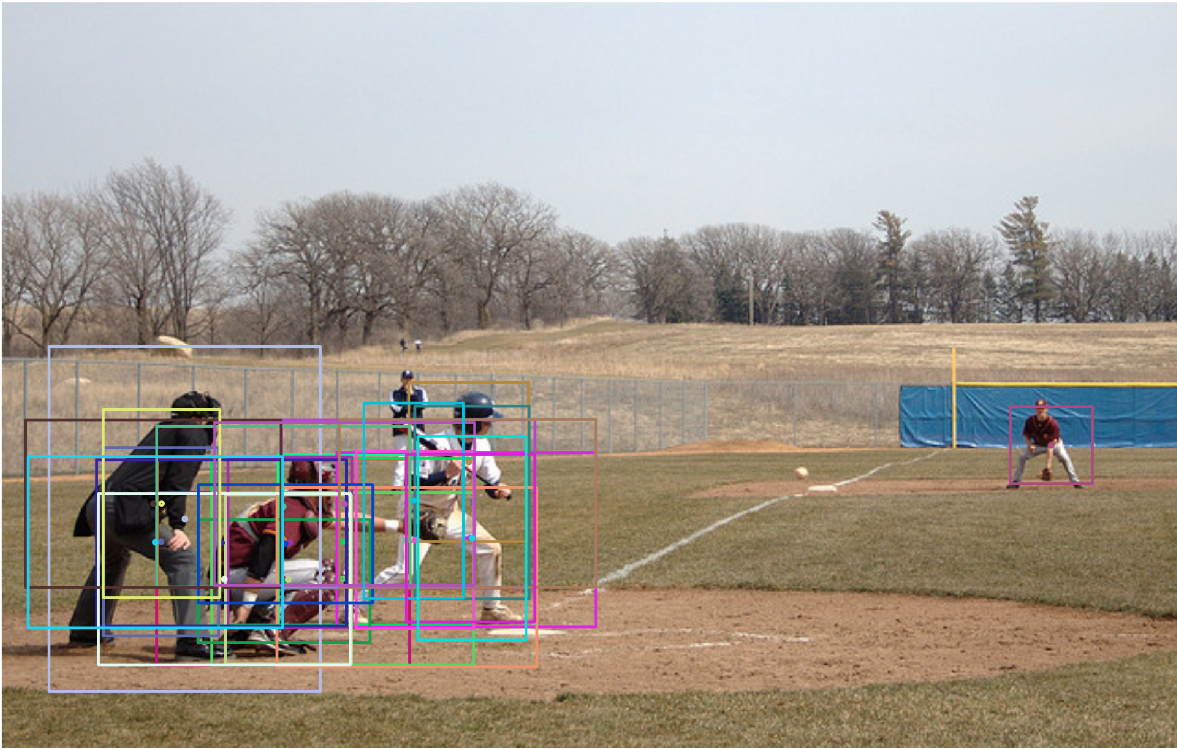}
    \caption{(left) A single forward pass through SSD \cite{Liu16} yields 9 individual object detections $D_i$. (center) 42 forward passes with Dropout Sampling result a total of 393 detections $D_i$. (right) These individual detections can be grouped according to their $\iou$ score into 29 observations $\cO_j$.}
    \label{fig:SSD_passes}
\end{figure*}

A single forward pass through a sampled object detection network with weights $\widetilde\vW$ yields a set of individual detections, each consisting of bounding box coordinates $\vb$ and a softmax score vector $\vs$. We denote these detections as $D_i=\{\vs_i, \vb_i\}$. Multiple forward passes yield a larger set $\fD = \{D_1, \dots, D_n\}$ of $n$ such individual detections $D_i$. Notice that many of these detections $D_i$ will overlap significantly as they correspond to objects that are detected in every single forward pass. This is illustrated in Fig.~\ref{fig:SSD_passes}.

Detections from the set $\fD$ with high mutual intersection-over-union scores ($\iou$) will be partitioned into \emph{observations} using a Union-Find data structure. We define an observation $\cO_i$ as a set of detections with high mutual bounding box $\iou$:  
\begin{equation}
    \cO_i = \cup D_i \;\;\; \text{s.t.} \; \iou(D_j,D_k) \ge 0.95 \;\; \forall D_j, D_k \in \cO_i    
\end{equation}
The threshold of $0.95$ has been determined empirically. Smaller thresholds (e.g. 0.8 in our experiments) tend to group too many overlapping detections into one observation in cluttered scenes, often falsely grouping detections on different ground truth objects into one observation. The selected threshold of 0.95 is conservative, resulting in several observations per object. We found that this conservative partitioning strategy is a better choice, as it is easier to fuse observations at later stages in the processing pipeline through data association techniques than it is to re-separate wrongly combined detections.

\subsection{Extracting Label Probabilities and Uncertainty}
When performing dropout sampling with multiple forward passes and partitioning of individual detections into observations as described above, we obtain a set of score vectors for every observation. 
Following \eqref{eq:do_sample} we can now approximate the vector of class probabilities $\vq_i$ by averaging all score vectors $\vs_j$ in an observation $\cO_i$.
\begin{equation}
    \vq_i \approx \bar\vs_i = \frac{1}{n}\sum_{j=1}^n \vs_j \;\;\;\; \forall D_j=\{\vs_j, \vb_j\} \in \cO_i
\end{equation}
This gives us an approximation of the probability of the class label $y_i$ for a detected object in image $\cI$ given the training data $\vT$, which follows a Categorical distribution parameterized by $\vq_i$ and the number of classes $k$:
\begin{equation}
    p(y_i | \cI, \vT) \sim \cat{k}{\vq_i}
\end{equation}
The entropy $ H(\vq_i) = -\sum_j q_{ij} \cdot \log q_{ij}$
measures the \emph{label uncertainty} of the detector for a particular observation. If $\vq_i$ is a uniform distribution, expressing maximum uncertainty, the Entropy will be high. Conversely, if the detector is very certain and puts most of its probability mass into a single class, resulting in a very ``peaky'' distribution, the entropy will be low.

\subsection{Extracting Location Probability and Spatial Uncertainty}
While the averaged Softmax scores approximate the label distribution $\vq_i$, we can approximate the distribution over the bounding box coordinates for every observation in the same way: by averaging over the bounding box vectors $\vb_j$ of all detections $D_j$ belonging to an observation $\cO_i$:
\begin{equation}
    \bar{\vb_i} = \frac{1}{n}\sum_{j=1}^n \vb_j \;\;\;\; \forall D_j=\{\vs_j, \vb_j\} \in \cO_i
\end{equation}
The uncertainty in these bounding box coordinates is captured by the covariance matrix over all $\vb_j$. While we do not use this expression of \emph{spatial} uncertainty in this paper, it can be of use for future applications such as utilizing the bounding box detections as landmark parametrizations in object-based SLAM \cite{Suenderhauf2017Quadrics}.

\subsection{Using Dropout Sampling to Improve Object Detection Performance in Open-Set Conditions}
\label{sec:hypothesis}
The described dropout sampling technique for object detection allows us to estimate the \emph{uncertainty} of the detector in the label classification for every observation $\cO_i$ by assessing the Entropy $H(\vq_i$). In open-set conditions, we would expect the label uncertainty to be higher for detections falsely generated on open-set objects (i.e. \emph{unknown} object classes not contained in the training data). A threshold on the Entropy $H(\vq_i)$ allows us to identify and reject detections of such unknown objects. 

While the same Entropy test could be applied to the Entropy of a single Softmax score vector $H(\vs)$ from the vanilla, non-Bayesian object detector network, we would expect that since $\vq_i$ is a better approximation to the true class probability distribution than $\vs$, using $H(\vq_i)$ as a measure of uncertainty is superior over $H(\vs)$.

This allows us to formulate the central {\bf Hypothesis} of our paper: \emph{Dropout variational inference improves the object detection performance under open-set conditions compared to a non-Bayesian detection network.}  
The following two sections describe the experiments we conducted to verify or falsify this hypothesis and present our findings.

\section{Evaluation Metrics}\label{sec:metrics}

We evaluate the object detection performance in open-set conditions with three metrics: (1)~open-set error, (2)~precision and (3)~recall. Recall describes how well a detector identifies \emph{known} objects, open-set error describes how robust an object detector is with respect to \emph{unknown} objects and precision describes how well a detector classifies \emph{known} and \emph{unknown} objects. An ideal object detector would achieve a recall of 100\% (it detects \emph{all} known objects), precision of 100\% (\emph{all} detections are classified correctly as the true \emph{known} class or as \emph{unknown}), and an open-set error of 0 (no \emph{unknown} objects were detected and misclassified as a \emph{known} class).

\subsection{Precision and Recall}
We define precision and recall by arranging all observations in a scene into true positives (TP) and false positives (FP). Ground truth objects that are not detected are counted as false negatives (FN). 

Let $\Omega = \{\cO_1, \dots \cO_n\}$ be the set of \emph{all} object observations in a scene after the partitioning step described in Section \ref{sec:partition}. We assess the label uncertainty by comparing the Entropy $H(\vq_i)$ with a threshold $\theta$ and reject a detection if $H(\vq_i) > \theta$. The rejected detections exhibit high label uncertainty and are likely to correspond to observations of unknown objects.

For every observation $\cO_i$ that passes this Entropy test, we find the set of overlapping ground truth objects with an $\iou$ of at least $0.5$. This is an established minimum requirement for coupling a detection with a ground truth object \cite{Lin14}. If the winning label for the observation matches any of the matched objects, we count the observation as true positive, otherwise as false positive. 

Should there be no ground truth object with an $\iou \geq 0.5$ and the winning class label is not 0 (unknown), we also count $\cO_i$ as a false positive. This case corresponds to observations that passed the Entropy test, but were not generated by a known object.

Every ground truth object of a class known to the detector that was not associated with an observation (i.e. there is no $\cO_i$ with an $\iou \geq 0.5$ with that object) gets counted as a false negative, as the detector failed to detect the \emph{known} object.

Precision and recall are then defined as usual: $\text{precision} = \frac{|TP|}{|TP| + |FP|}$,  and $\text{recall} = \frac{|TP|}{|TP| + |FN|}$. Both can be combined into the F-score $F_1 = 2 \cdot \frac{\text{precision} \cdot  \text{recall}}{\text{precision} + \text{recall}}$.

\subsection{Absolute Open-Set Error}

We define absolute open-set error as the total number of observations that pass the Entropy test, fall on unknown objects (i.e. there are no overlapping ground truth objects with an $\iou \geq 0.5$ and a known true class label) and do not have a winning class label of 'unknown'.

In the ideal case, all observations are of \emph{known} objects, i.e. objects from the training set. In this scenario the open-set error is 0.

\subsection{Datasets Used in the Evaluation}
Our evaluation is based on two datasets: SceneNet RGB-D~\cite{McCormac16a}, a huge dataset of rendered scenes, and the QUT Campus dataset, a smaller real-world dataset captured by our robot in a variety of indoor and outdoor environments on our campus~\cite{Suenderhauf15c}.

\paragraph{SceneNet RGB-D}
The SceneNet RGB-D validation set contains photo-realistic images of 1000 differing indoor scenes \cite{McCormac16a}. These scenes contain 182 differing objects, of which 100 are unknown classes for a network trained on COCO. Instance images from the dataset contain pixel segmentations of each object and can be used to obtain ground truth locations and classifications. A bounding box was generated for each object by extracting it's minimum and maximum x and y pixel locations in the instance image. The instance ID for that object was then mapped to a WordNet ID (wnid) via the dataset's trajectories. A map was created to convert each COCO class to all corresponding wnids in the dataset. As COCO classes are more generic in nature, several wnids were often mapped to a single COCO class, i.e. 'rocking chair', 'swivel chair' and 'arm chair' were mapped to the COCO class 'chair'.

\paragraph{QUT Campus Dataset} 
This dataset was collected using a mobile robot 
across nine different and versatile environments on our
campus while recording stream of images. The traversed environments are an office, a corridor, the underground parking garage, a small supermarket, a food court, a cafe, a general outdoor campus environment, a lecture theater and the lobby of one of the university’s main buildings. More details about the dataset can be found in~\cite{Suenderhauf15c}. Detections were evaluated by manual visual inspection. 

\subsection{Evaluation Protocol and Compared Object Detectors}
We base our evaluation on the SSD architecture~\cite{Liu16} and compare the performance of three variants:
\begin{itemize}
    \item Vanilla SSD, i.e. the default configuration of SSD as proposed in~\cite{Liu16}, without any Entropy thresholding
    \item SSD with Entropy thresholding, i.e. using the Entropy of the Vanilla SSD Softmax scores $H(\vs)$ to estimate uncertainty and reject detections
    \item Bayesian SSD, i.e. SSD with Dropout Sampling and using the Entropy of the averaged Softmax scores $H(\vq)$ to estimate uncertainty and reject detections
\end{itemize}

Two key parameters of Bayesian SSD are the number of forward passes through the network and the minimum number of detections required per observation. More forward passes is expected to improve recall performance at the cost of processing time. Bayesian SSD was tested for 10, 20, 30 and 42 forward passes through the network to verify this. Given that Bayesian SSD relies on partitioning and averaging across individual detections, it can be expected that observations containing more individual detections will provide more robust uncertainty estimates. Minimum requirements of 1, 3, 5 and 10 detections per observation were evaluated for 42 forward passes. 

We varied the Entropy threshold $\theta$ between 0.1 and 2.5 and calculated precision, recall, and open-set error for every $\theta$. Each network was fine-tuned on the COCO dataset. From each scene of the SceneNet RGB-D validation dataset, we tested 30 images, resulting in a total of 30000 test images.
A sample of 75 images were tested from the QUT Campus dataset across 11 scenes with absolute true detections and error recorded.

\section{Results and Interpretation}\label{sec:results}
\subsection{Summary}

Our experiments confirmed the hypothesis formulated in Section \ref{sec:hypothesis}: The Bayesian SSD detector utilizing Dropout Sampling as an approximation to full Bayesian inference improved the object detection performance in precision and recall while reducing the open-set error in open-set conditions.

We will explain our findings in detail in this section, discussing the results on both datasets as well as the influence of the hyper parameters for the number of forward passes and the required minimum detections per observation.

\begin{table}[bt]
\centering
\caption{Performance Comparison on SceneNet RGB-D at maximum $F_1$ score \cite{McCormac16a}}
\label{tab:performance_max}
\scalebox{0.84}{%
\begin{tabular}{@{}rcccc@{}}
 \toprule
 Forward  & max.        & abs OSE & Recall & Precision \\
 Passes   & $F_1$ Score & \multicolumn{3}{c}{at max $F_1$ point} \\ \midrule
 vanilla SSD & 0.220 & 18331 & 0.165 & 0.328\\
 SSD with Entropy test & 0.227 & \bf 12638 & 0.160 & \bf 0.392 \\\midrule
 Bayesian SSD \hspace{0.5cm} 10 & 0.270 & 20991 & 0.214 & 0.364 \\
 20 & 0.292 & 24922 & 0.244 & 0.364 \\
 30 & 0.301 & 28431 &0.261 & 0.355 \\ 
 42 &  \bf 0.309 & 32034 & \bf 0.278 & 0.347 \\
 \bottomrule
\end{tabular}}
\end{table}

\begin{table}[bt]
\centering
\caption{Performance Comparison on SceneNet RGB-D at Vanilla SSD reference scores \cite{McCormac16a}}
\label{tab:performance_reference}
\scalebox{0.84}{%
\begin{tabular}{@{}rcc@{}}
 \toprule
 Forward  &  $F_1$ Score at & abs OSE at  \\
 Passes   & reference OSE & reference $F_1$ Score \\ \midrule
 vanilla SSD (reference) & 0.220 & 18,331 \\\midrule
 Bayesian SSD \hspace{0.5cm} 10 &  0.269 & \bf 8,225 \\
 20 &  0.284 & 8,313  \\
 30 &  \bf 0.286 & 9,003\\ 
 42 &  0.285 & 9,256 \\
 \bottomrule
\end{tabular}}
\end{table}

\subsection{SceneNet RGB-D}
\begin{figure}[t]
    \centering
    \includegraphics[width=0.8\linewidth]{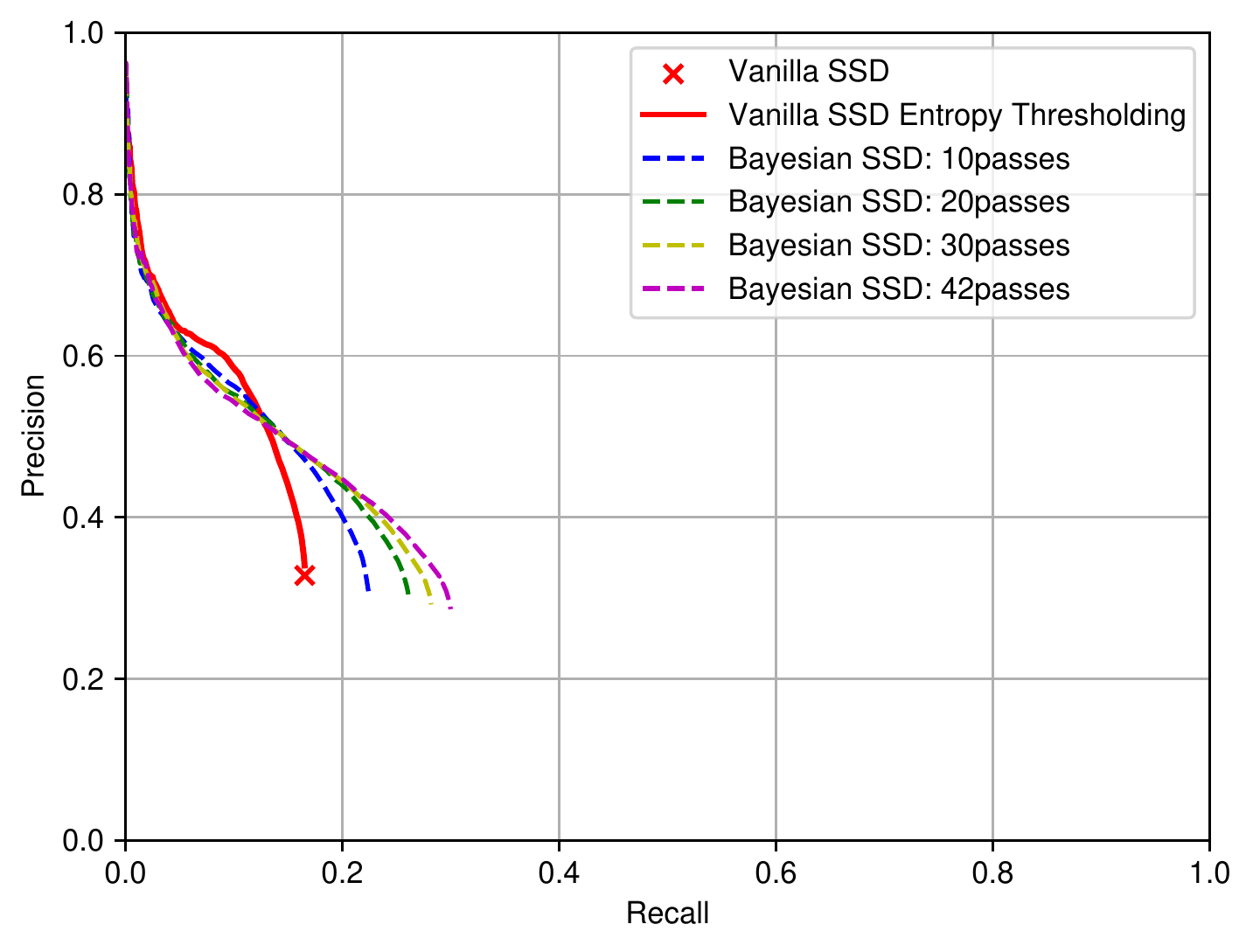}
    \caption{Precision-recall curves for each network tested on SceneNet RGBD when thresholding softmax entropy.}
    \label{fig:prec_rec}
\end{figure}

 As shown in Table \ref{tab:performance_max} and Figure \ref{fig:prec_rec}, Bayesian SSD is able to achieve greater precision and recall scores than the vanilla SSD. At the same precision performance ($32.8\%$) as the vanilla SSD, Bayesian SSD demonstrates a 12.3\% increase in recall; similarly, for the same recall score ($16.5\%$), Bayesian SSD demonstrates a 15.1\% increase in precision. While the SSD with Entropy thresholding network has a higher precision for some low recall levels, overall, Bayesian SSD is also shown to outperform this approach. This suggests that Bayesian SSD produces a more reliable uncertainty estimate for object classification; as such, it is able to make more informed decisions to reject incorrect classifications. A network utilizing Bayesian SSD is also able to achieve a considerably higher maximum recall. As expected, collecting detections from multiple forward passes allows Bayesian SSD to have a greater chance of detecting objects that may be overlooked in a single forward pass.

\begin{figure}[t]
    \centering
    \includegraphics[width=0.8\linewidth]{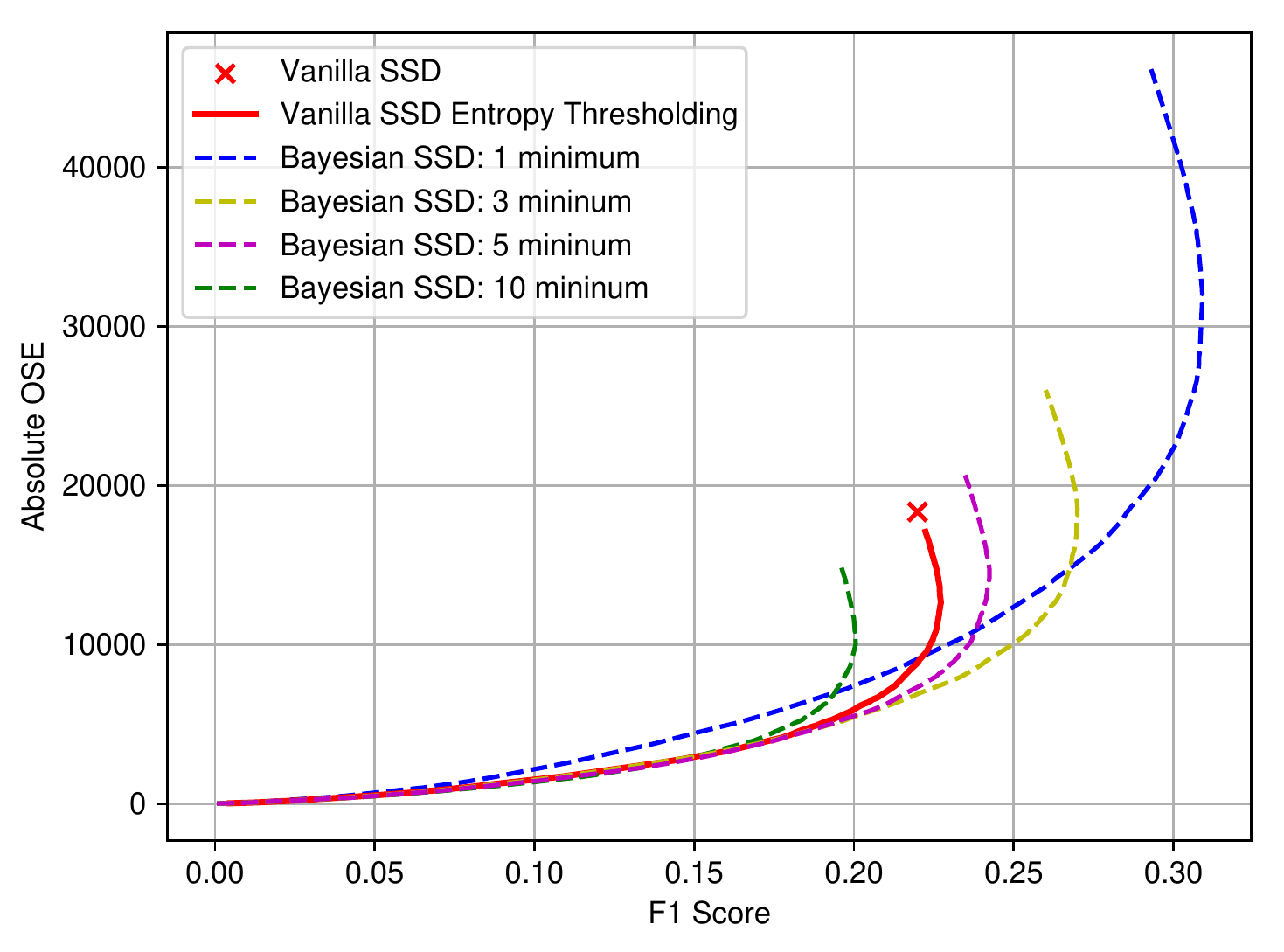}
    \caption{F1 score versus open-set error for various minimum detection requirements. Perfect performance is an F1 score of 1 and an Absolute OSE of 0.}
    \label{fig:F1_min}
\end{figure}

The effect of Bayesian SSD on identification of open-set error is further explored in Figure \ref{fig:f1_ose}. These results show that the Bayesian SSD allows for a reduction in open-set error in comparison to vanilla SSD. As can be seen in Table \ref{tab:performance_reference}, when choosing the performance of the vanilla SSD as a reference point (indicated by the red cross in Fig. \ref{fig:F1_min}) the Bayesian SSD allows a decrease the open-set error (OSE) while retaining the $F_1$ score. Alternatively the $F_1$ can be substantially improved while keeping the OSE at the reference level.
This further suggests that Bayesian SSD provides a reliable uncertainty measure for identifying incorrect detections of unknown classes, as well as incorrect classifications of known objects.
 
\subsection{Forward Passes} 
\begin{figure}[t]
    \centering
    \includegraphics[width=0.8\linewidth]{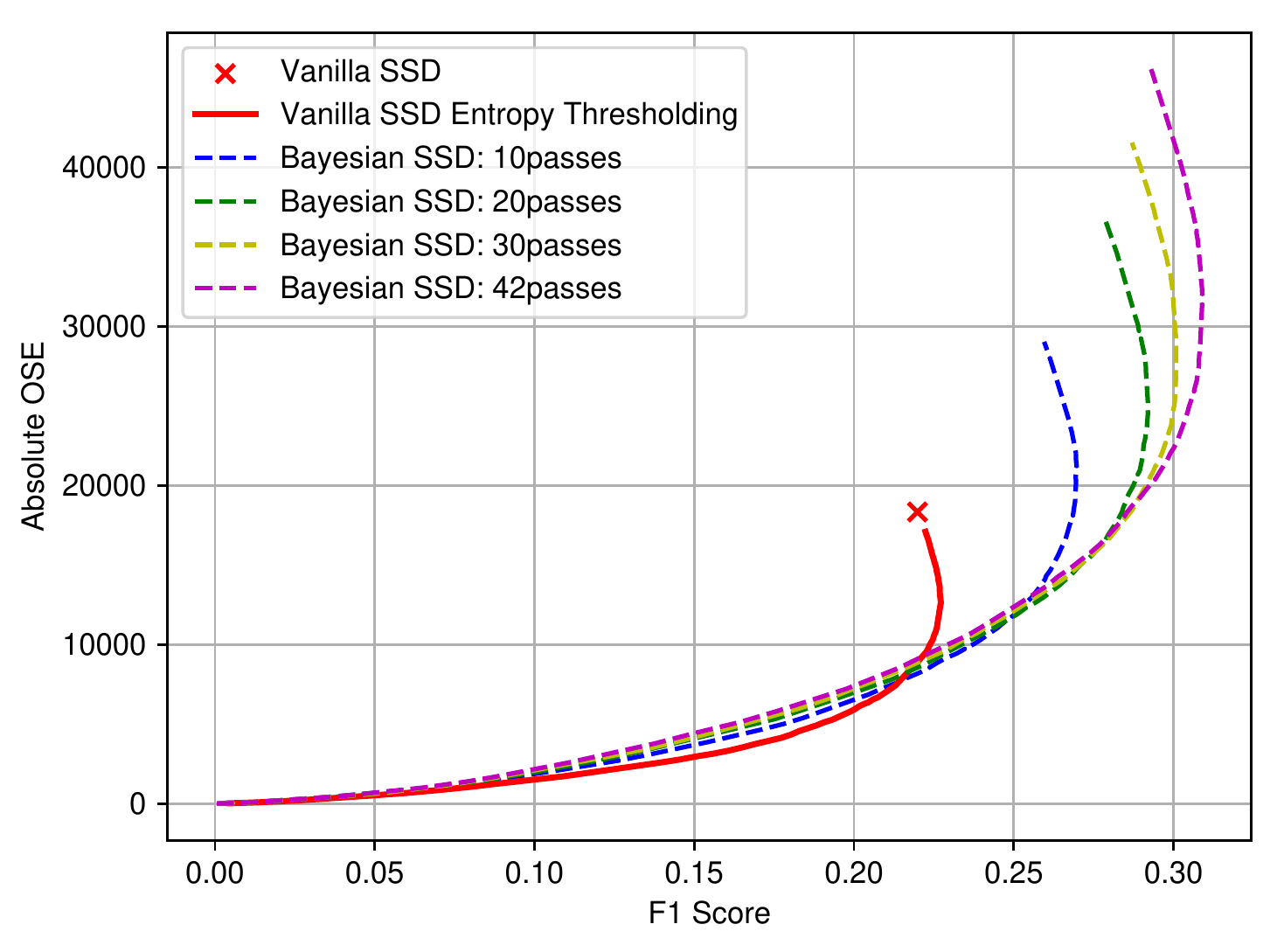}
    \caption{F1 score versus open-set error for each network. Perfect performance is an F1 score of 1 and an Absolute OSE of 0.}
    \label{fig:f1_ose}
\end{figure}

As can be seen in Figure \ref{fig:f1_ose}, as few as 10 forward passes is able to maintain the vanilla SSD reference $F_1$ score and reduce open-set error comparably to greater numbers of passes. However, at least 20 forward passes are needed to maximize $F_1$ score for the vanilla SSD reference open-set error. Beyond the reference OSE point, more forward passes achieve slightly higher $F_1$ scores, but at the cost of a large increase in open-set error. As the open-set error increases, recall of the system increases while precision decreases. At very high open-set error levels, precision is low enough to decrement the $F_1$ score despite the high recall; this causes the backward bending trend as shown in Figure \ref{fig:f1_ose}. Depending on the performance requirements of a detection system, fewer forward passes may be suitable, thus allowing for reduced computation. One forward pass of an image takes 0.05 seconds with the current model, which currently involves passing an image through the entire network. In future, computation could be reduced by only sampling over the post-dropout layers (inclusive of the dropout layers) component of the network, as all computation prior to this point is not stochastic.

\begin{figure}[t]
    \centering
    \includegraphics[width=0.8\linewidth]{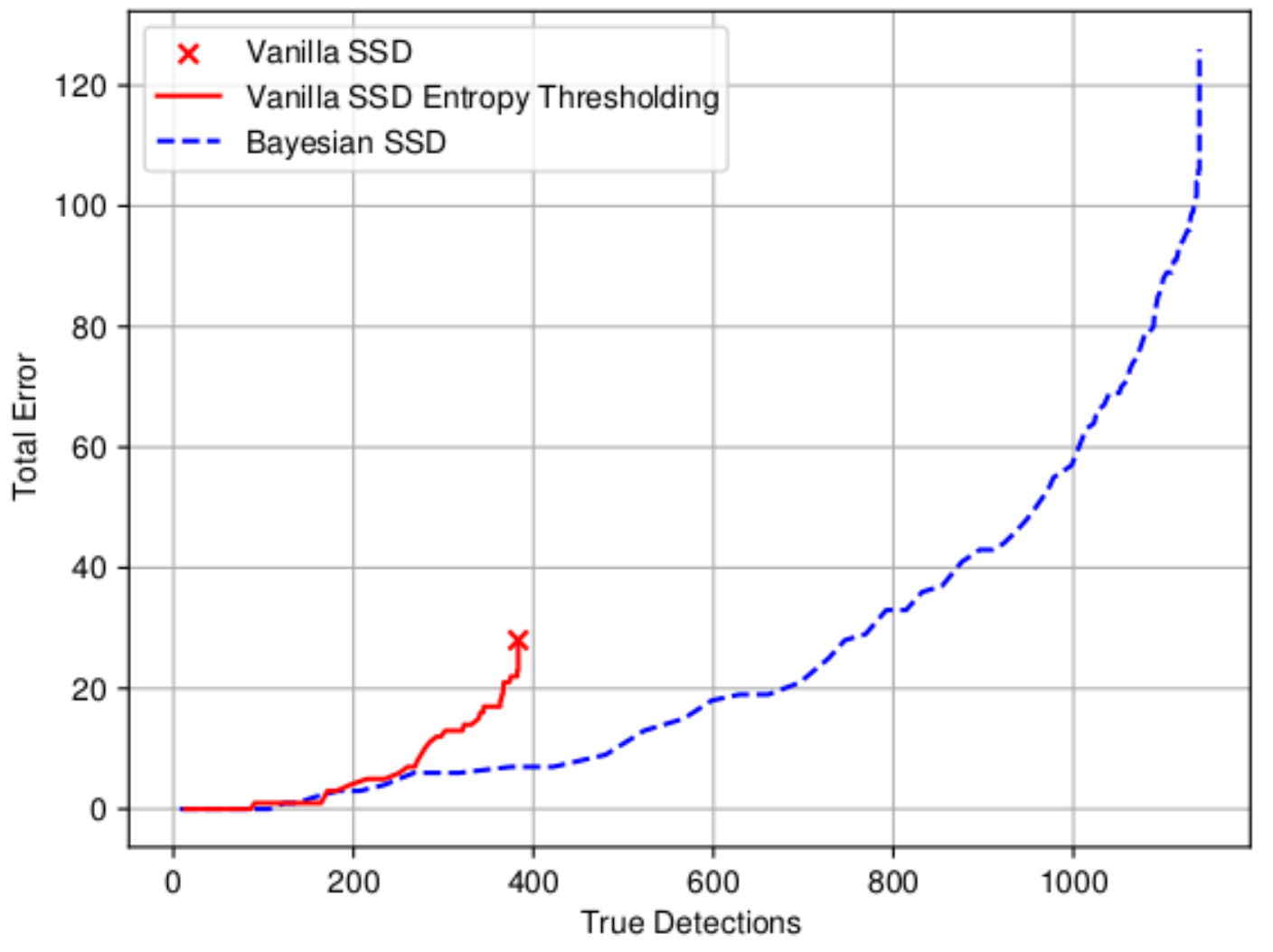}
    \caption{True detections versus total error for QUT Campus dataset.}
    \label{fig:realworld}
\end{figure}
\subsection{Minimum Detection} 
 As shown in Figure \ref{fig:F1_min}, requiring at least 3 detections per observation provides a marginally lower open-set error for each F1 score. This effect is equivalent across all minimum detection levels greater than 1. As a consequence of this requirement, the maximum F1 score is also reduced. As in the case of 10 minimum detections, this can result in Bayesian SSD being outperformed by vanilla SSD. This supports the theory that Bayesian SSD relies upon having multiple detections per observation, but also suggests that the magnitude is inconsequential. Therefore, in most circumstances, a low minimum detections requirement (if any) is ideal.

\begin{figure*}[t]
    \centering
    \includegraphics[width=0.3\linewidth]{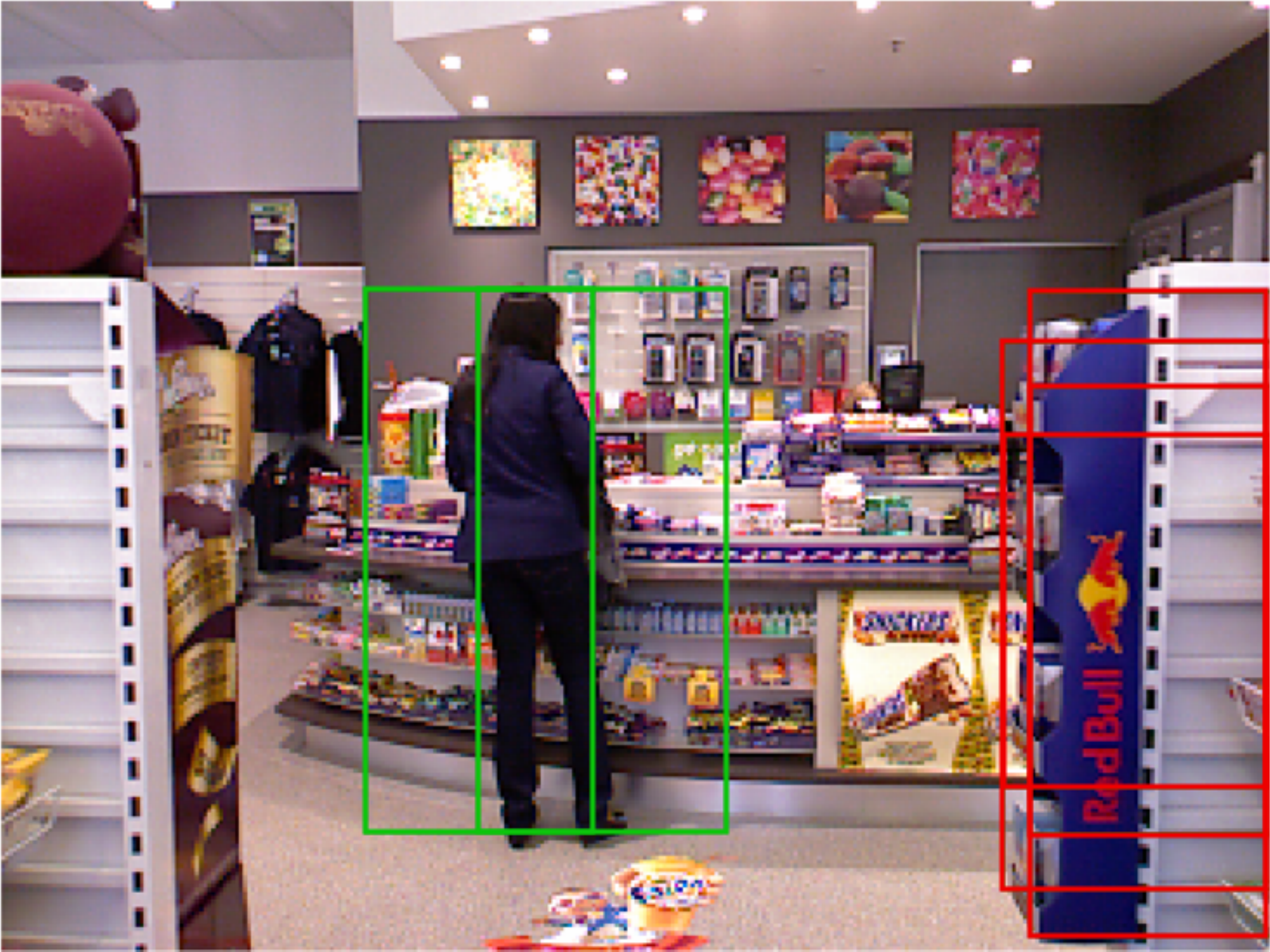}
    \includegraphics[width=0.3\linewidth]{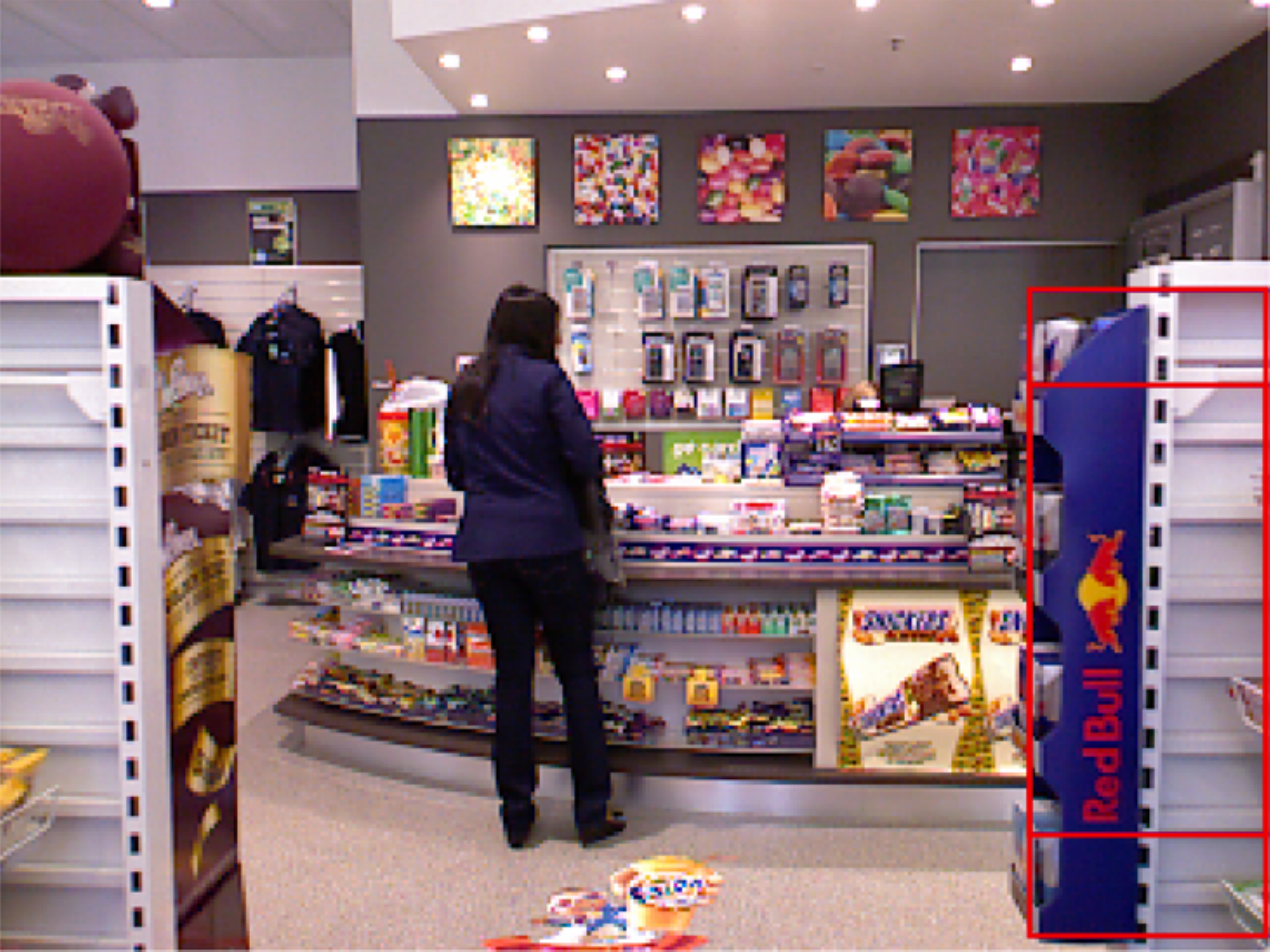}
    \includegraphics[width=0.3\linewidth]{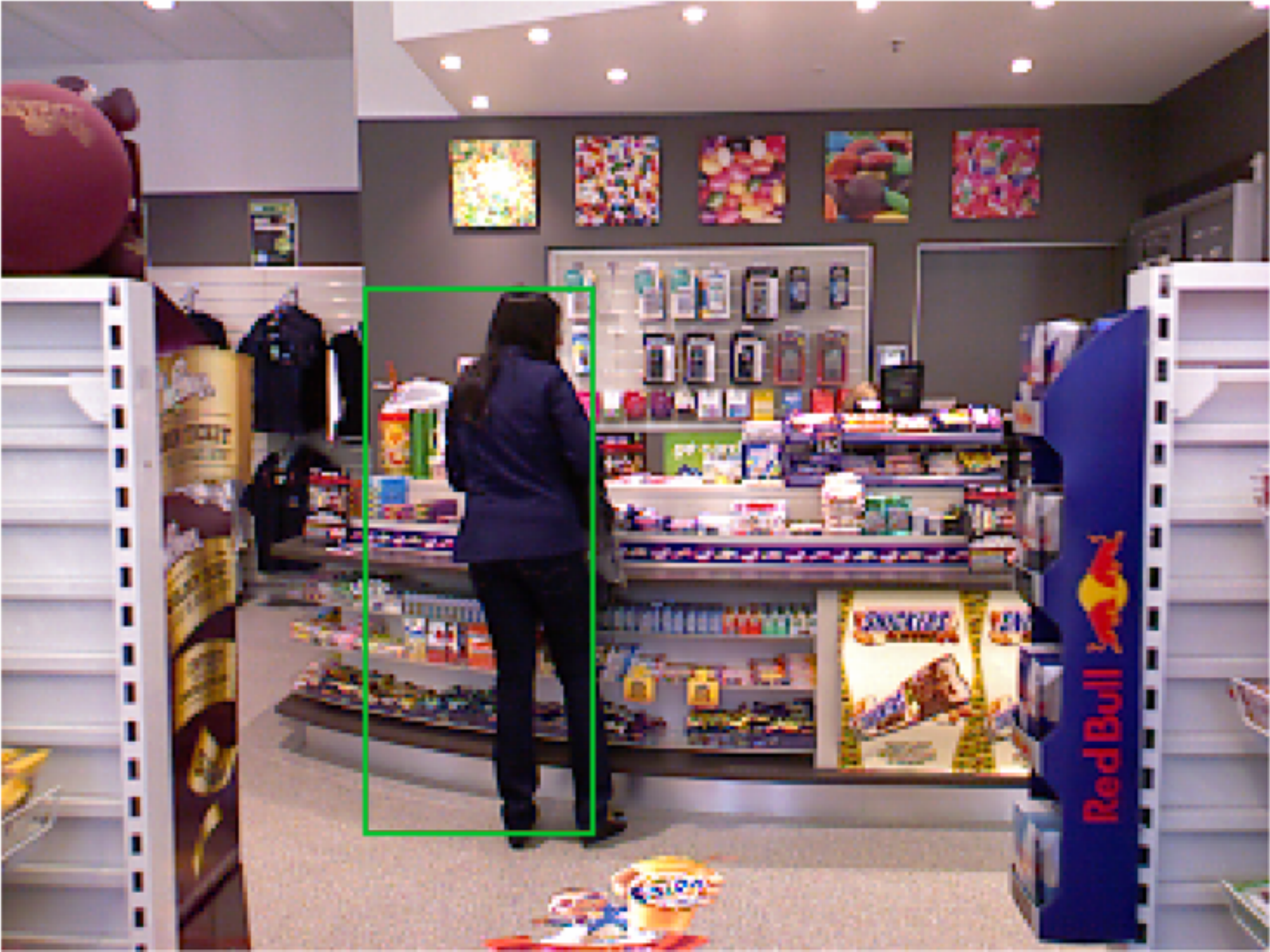}
    \caption{True detections are shown in green and open-set errors are shown in red. Vanilla SSD (left) detecting two true detections of 'person' and four open-set errors of 'refrigerator'. Vanilla SSD with thresholding (center) detecting two open-set errors of 'refrigerator'. Bayesian SSD (right) detecting one true detection of 'person'. Entropy thresholding at 0.64.}
    \label{fig:eg1}
\end{figure*}

\subsection{Real World Dataset}
For the QUT Campus dataset, the Bayesian SSD is able to reduce the total error per true detection. This can be seen in Figure \ref{fig:realworld}, where at the reference point for the vanilla SSD with no entropy thresholding, Bayesian SSD has substantially reduced the total error by a margin of 21 (consisting of open-set error and incorrect classifications of known objects). Additionally, for the same total error, Bayesian SSD achieves a greater number of true detections by a margin of 363. While this may be due to multiple detections per object, it can also be inferred that this partially represents the superior recall performance of Bayesian SSD.

Examples of each network's performance on an image from the dataset are shown in Figure \ref{fig:eg1}. For this image, an entropy threshold of 0.64 was applied. As can be seen, the vanilla SSD makes correct detections of a person as well as several open-set errors (an unknown object, a drink shelf, is detected four times as a 'refrigerator'). When applying entropy thresholding to the vanilla SSD, all true detections are discarded while most of the open-set error is sustained. In contrast, Bayesian SSD is able to utilize its uncertainty to preserve a true detection of the person while eliminating all open-set error.

\section{Conclusions and Future Work}\label{sec:conc}
We showed that Dropout Sampling is a practical way of performing object detection with an approximated Bayesian network. We verified the central hypothesis of our paper that Dropout Sampling allows to extract better label uncertainty information and thereby helps to improve the performance of object detection in the open-set conditions that are ubiquitous for mobile robots.

A promising direction for future work is to exploit the \emph{spatial} uncertainty contained in the covariance matrix over the bounding box coordinates for a group of detections. This information could be propagated through a object-based SLAM system to gain a better estimate of the 6-DOF object pose.

\bibliographystyle{ieeetr}
\bibliography{bibfile,review}

\end{document}

%% file: mathstuff.tex
\newcommand{\vect}[1]{\mathbf{ #1}}

\newcommand{\normal}[2]{\mathcal{N}\left(#1, #2\right)}

\newcommand{\cat}[2]{\text{Cat}\left(#1, #2\right)}

\def\iou{\mathop{\mathrm {IoU}}}

\newcommand{\vI}{\vect{I}}

\newcommand{\vT}{\vect{T}}

\newcommand{\vW}{\vect{W}}

\newcommand{\vb}{\vect{b}}

\newcommand{\vq}{\vect{q}}

\newcommand{\vs}{\vect{s}}

\newcommand{\cI}{\mathcal{I}}

\newcommand{\cO}{\mathcal{O}}

\newcommand{\fD}{\mathfrak{D}}

%% file: dropout_sampling.bbl
\begin{thebibliography}{10}

\bibitem{Liu16}
W.~Liu, D.~Anguelov, D.~Erhan, C.~Szegedy, S.~Reed, C.-Y. Fu, and A.~C. Berg,
  ``{SSD: Single shot multibox detector},'' in {\em European conference on
  computer vision}, pp.~21--37, Springer, 2016.

\bibitem{YOLO9000}
J.~Redmon and A.~Farhadi, ``Yolo9000: Better, faster, stronger,'' in {\em 2017
  IEEE Conference on Computer Vision and Pattern Recognition (CVPR)},
  pp.~6517--6525, July 2017.

\bibitem{Ren15}
S.~Ren, K.~He, R.~Girshick, and J.~Sun, ``{Faster R-CNN: Towards real-time
  object detection with region proposal networks},'' in {\em {Advances in
  Neural Information Processing Systems (NIPS)}}, pp.~91--99, 2015.

\bibitem{scheirer2013toward}
W.~J. Scheirer, A.~de~Rezende~Rocha, A.~Sapkota, and T.~E. Boult, ``Toward open
  set recognition,'' {\em IEEE Transactions on Pattern Analysis and Machine
  Intelligence}, vol.~35, no.~7, pp.~1757--1772, 2013.

\bibitem{scheirer2014probability}
W.~J. Scheirer, L.~P. Jain, and T.~E. Boult, ``Probability models for open set
  recognition,'' {\em IEEE transactions on pattern analysis and machine
  intelligence}, vol.~36, no.~11, pp.~2317--2324, 2014.

\bibitem{Torralba11}
A.~Torralba and A.~A. Efros, ``Unbiased look at dataset bias,'' in {\em
  Computer Vision and Pattern Recognition (CVPR), 2011 IEEE Conference on},
  pp.~1521--1528, IEEE, 2011.

\bibitem{gal2016dropout}
Y.~Gal and Z.~Ghahramani, ``Dropout as a bayesian approximation: Representing
  model uncertainty in deep learning,'' in {\em International Conference on
  Machine Learning (ICML)}, pp.~1050--1059, 2016.

\bibitem{kendall2016bayesian}
A.~Kendall, V.~Badrinarayanan, and R.~Cipolla, ``Bayesian segnet: Model
  uncertainty in deep convolutional encoder-decoder architectures for scene
  understanding,'' {\em arXiv preprint arXiv:1511.02680}, 2016.

\bibitem{kendall2016modelling}
A.~Kendall and R.~Cipolla, ``Modelling uncertainty in deep learning for camera
  relocalization,'' in {\em Robotics and Automation (ICRA), 2016 IEEE
  International Conference on}, pp.~4762--4769, IEEE, 2016.

\bibitem{Lin14}
T.-Y. Lin, M.~Maire, S.~Belongie, J.~Hays, P.~Perona, D.~Ramanan,
  P.~Doll{\'a}r, and C.~L. Zitnick, ``{Microsoft COCO: Common objects in
  context},'' in {\em European Conference on Computer Vision (ECCV)},
  pp.~740--755, Springer, 2014.

\bibitem{Russakovsky15}
O.~Russakovsky, J.~Deng, H.~Su, J.~Krause, S.~Satheesh, S.~Ma, Z.~Huang,
  A.~Karpathy, A.~Khosla, M.~Bernstein, A.~C. Berg, and L.~Fei-Fei, ``{ImageNet
  Large Scale Visual Recognition Challenge},'' {\em International Journal of
  Computer Vision}, vol.~115, no.~3, pp.~211--252, 2015.

\bibitem{Girshick14}
R.~Girshick, J.~Donahue, T.~Darrell, and J.~Malik, ``Rich feature hierarchies
  for accurate object detection and semantic segmentation,'' in {\em
  Proceedings of the IEEE Conference on Computer Vision and Pattern Recognition
  ({CVPR})}, 2014.

\bibitem{Krizhevsky12}
A.~Krizhevsky, I.~Sutskever, and G.~E. Hinton, ``Imagenet classification with
  deep convolutional neural networks,'' in {\em Advances in Neural Information
  Processing Systems 25}, 2012.

\bibitem{bendale2016towards}
A.~Bendale and T.~E. Boult, ``Towards open set deep networks,'' in {\em
  Proceedings of the IEEE Conference on Computer Vision and Pattern
  Recognition}, pp.~1563--1572, 2016.

\bibitem{rudd2017extreme}
E.~M. Rudd, L.~P. Jain, W.~J. Scheirer, and T.~E. Boult, ``The extreme value
  machine,'' {\em IEEE Transactions on Pattern Analysis and Machine
  Intelligence}, 2017.

\bibitem{mackay1992practical}
D.~J. MacKay, ``A practical bayesian framework for backpropagation networks,''
  {\em Neural computation}, vol.~4, no.~3, pp.~448--472, 1992.

\bibitem{neal1995bayesian}
R.~M. Neal, {\em Bayesian learning for neural networks}.
\newblock PhD thesis, University of Toronto, 1995.

\bibitem{paisley2012variational}
J.~Paisley, D.~Blei, and M.~Jordan, ``Variational bayesian inference with
  stochastic search,'' in {\em Proceedings of International Conference on
  Machine Learning (ICML)}, 2012.

\bibitem{kingma2013auto}
D.~P. Kingma and M.~Welling, ``Auto-encoding variational bayes,'' in {\em
  Proceedings of International Conference on Learning Representations (ICLR)},
  2014.

\bibitem{rezende2014stochastic}
D.~J. Rezende, S.~Mohamed, and D.~Wierstra, ``Stochastic backpropagation and
  approximate inference in deep generative models,'' in {\em Proceedings of the
  International Conference on Machine Learning (ICML)}, 2014.

\bibitem{titsias2014doubly}
M.~Titsias and M.~L{\'a}zaro-Gredilla, ``Doubly stochastic variational bayes
  for non-conjugate inference,'' in {\em Proceedings of the 31st International
  Conference on Machine Learning (ICML-14)}, pp.~1971--1979, 2014.

\bibitem{hoffman2013stochastic}
M.~D. Hoffman, D.~M. Blei, C.~Wang, and J.~Paisley, ``Stochastic variational
  inference,'' {\em The Journal of Machine Learning Research}, vol.~14, no.~1,
  pp.~1303--1347, 2013.

\bibitem{kendall2017uncertainties}
A.~Kendall and Y.~Gal, ``What uncertainties do we need in bayesian deep
  learning for computer vision?,'' in {\em Advances in Neural Information
  Processing Systems (NIPS)}, pp.~5580--5590, 2017.

\bibitem{simonyan2014SSD}
K.~Simonyan and A.~Zisserman, ``Very deep convolutional networks for
  large-scale image recognition,'' in {\em Proceedings of International
  Conference on Learning Representations (ICLR)}, 2015.

\bibitem{Suenderhauf2017Quadrics}
N.~S\"underhauf and M.~Milford, ``Dual quadrics from object detection bounding
  boxes as landmark representations in slam,'' {\em arXiv preprint
  arXiv:1708.00965}, 2017.

\bibitem{McCormac16a}
J.~McCormac, A.~Handa, S.~Leutenegger, and A.~J. Davison, ``Scenenet rgb-d: Can
  5m synthetic images beat generic imagenet pre-training on indoor
  segmentation?,'' in {\em 2017 IEEE International Conference on Computer
  Vision (ICCV)}, pp.~2697--2706, Oct 2017.

\bibitem{Suenderhauf15c}
N.~S\"underhauf, F.~Dayoub, S.~McMahon, B.~Talbot, R.~Schulz, P.~Corke,
  G.~W.~B. Upcroft, and M.~Milford, ``{Place Categorization and Semantic
  Mapping on a Mobile Robot},'' in {\em 2016 IEEE International Conference on
  Robotics and Automation (ICRA)}, 2016.

\end{thebibliography}
